\documentclass[final]{cvpr}

\usepackage{times}
\usepackage{epsfig}
\usepackage{graphicx}
\usepackage{amsmath}
\usepackage{amssymb}
\usepackage{bm}
\usepackage{mathrsfs}
\usepackage{multirow}
\usepackage{booktabs}
\usepackage{appendix}

\usepackage[pagebackref=true,breaklinks=true,colorlinks,bookmarks=false]{hyperref}

\def\cvprPaperID{420} 
\def\confYear{CVPR 2021}
\newcommand{\zongxin}[1]{{#1}}

\begin{document}

\title{DSC-PoseNet: Learning 6DoF Object Pose Estimation via 
Dual-scale Consistency}

\author{Zongxin Yang$^{1,2}$,
Xin Yu$^{2}$,
Yi Yang$^{2*}$\\
$^{1}$ Baidu Research $^{2}$ ReLER, University of Technology Sydney \\
{\tt\small zongxin.yang@student.uts.edu.au, \{xin.yu,yi.yang\}@uts.edu.au}
}

\maketitle
\pagestyle{empty}
\thispagestyle{empty}
\renewcommand{\thefootnote}{*}
\footnotetext{This work was done when Zongxin Yang interned at Baidu Research. Yi Yang is the corresponding author.}

\begin{abstract}
Compared to 2D object bounding-box labeling, it is very difficult for humans to annotate 3D object poses, especially when depth images of scenes are unavailable. 
This paper investigates whether we can estimate the object poses effectively when only RGB images and 2D object annotations are given. 
To this end, we present a two-step pose estimation framework to attain 6DoF object poses from 2D object bounding-boxes. 
In the first step, the framework learns to segment objects from real and synthetic data in a weakly-supervised fashion, and the segmentation masks will act as a prior for pose estimation.
In the second step, we design a dual-scale pose estimation network, namely DSC-PoseNet, to predict object poses by employing a differential renderer.
To be specific, our DSC-PoseNet firstly predicts object poses in the original image scale by comparing the segmentation masks and the rendered visible object masks. 
Then, we resize object regions to a fixed scale to estimate poses once again. In this fashion, we eliminate large scale variations and focus on rotation estimation, thus facilitating pose estimation. Moreover, we exploit the initial pose estimation to generate pseudo ground-truth to train our DSC-PoseNet in a self-supervised manner.
The estimation results in these two scales are ensembled as our final pose estimation.
Extensive experiments on widely-used benchmarks demonstrate that our method outperforms state-of-the-art models trained on synthetic data by a large margin and even is on par with several fully-supervised methods.
\end{abstract}

\section{Introduction}
The goal of object pose estimation is to estimate 6 degrees of freedom (DoF) of a given object, including 3D orientations and 3D translations, with respect to a camera. Considering an object may undergo various lighting changes and severe occlusions, estimating accurate poses from a single RGB image is quite challenging. 

Thanks to the recent advance of deep neural networks, many deep learning based pose estimation algorithms (\eg,~\cite{zakharovdpod,li2019CDPN}) have been proposed recently and achieved promising performance. However, due to the notorious data-hunger nature of deep networks, state-of-the-art methods usually require a large number of real images with 3D pose annotations for training.
Unlike 2D image labeling tasks, annotating 3D object poses is difficult, especially when only RGB images are provided to labelers. However, without sufficient accurately annotated 3D poses of RGB images, pose estimation methods may suffer severe performance degradation~\cite{wang2020self6d}.

Different from real images of which poses are very difficult to obtain, pose labels in synthetic data are easily accessible.
However, due to the domain gap between synthetic and real data, such as illumination variations and smoothed 3D reconstructed models, state-of-the-art fully-supervised pose estimation approaches would suffer drastic performance degradation when they are trained on synthesized images~\cite{wang2020self6d}. Therefore, we attempt to use minimal annotation efforts, \ie, 2D bounding boxes, to reduce the domain gap between real and synthetic images for pose estimation. We employ bounding-boxes to ensure objects of interest are correctly identified in images rather than highly relying on synthesized data's photorealism.

In this paper, we present a two-step object pose estimation framework to predict object poses by only leveraging easily obtained 2D annotations.
In the first step, we employ a weakly supervised segmentation method to distinguish object pixels from background ones. Specifically, we firstly initialize a segmentation network by training it on synthetic images. Motivated by~\cite{chen2020semi,zhang2020weakly}, we use the segmentation network to predict pseudo masks for the unlabeled real data and re-train the network on both synthetic and pseudo-labeled data. 
Since some pixels in the background might be recognized as foreground ones, we fully exploit our 2D bounding-boxes to remove outliers and thus facilitate learning segmentation. The segmentation results, in return, provide a strong prior for pose estimation.

In the second step, we present a dual-scale pose estimation network, dubbed DSC-PoseNet. 
Our DSC-PoseNet is firstly initialized by training on synthetic images. Then, we employ DSC-PoseNet to estimate object poses in the original image scale. 
DSC-PoseNet regresses an offset from each pixel to each object keypoint and employs an attention map to obtain the keypoint locations. We adopt differentiable EPnP~\cite{lepetit2009epnp} and a renderer~\cite{ravi2020pytorch3d} to achieve rendered object masks, and the intersection over union (IoU) between the segmentation results and rendered visible masks is employed to train our DSC-PoseNet on real images.
Once real image poses are attained, we resize object regions to a fixed scale and then project the keypoints computed by our estimated poses to the local regions as self-supervised signals to train DSC-PoseNet once again. By doing so, we improve the feature extraction ability of DSC-PoseNet and enforce the pose estimation consistency across different scales. 

As DSC-PoseNet outputs pose estimation results on two scales, we ensemble two-scale results to improve the estimation robustness. Extensive experiments on three popular widely-used datasets demonstrate that our method achieves superior performance compared to state-of-the-art RGB image based methods without real pose annotations.

Overall, our contributions are summarized as follows:
\begin{itemize}
\vspace{-0.5em}
\item We propose a weakly- and self-supervised learning based pose estimation framework to estimate object poses from single RGB images with easily obtained 2D bounding-box annotations.
\vspace{-0.5em}
\item We present a self-supervised dual-scale pose estimation network, named DSC-PoseNet. The DSC-PoseNet significantly alleviates the domain gap between synthetic and real data by constructing cross-scale self-supervision with a differentiable renderer. 
\vspace{-0.5em}    
\item To the best of our knowledge, our work is the first attempt to estimate 6DoF object poses from an RGB image without using 3D pose annotations and depth images \zongxin{during both training and testing phases}. Contrast results show that DSC-PoseNet outperforms RGB based competitors trained on synthetic data.
\end{itemize}

\section{Related Work}
Since our method predicts object poses from a single RGB image, we briefly review fully-supervised and self-supervised RGB image based pose estimation approaches.

\noindent\textbf{Conventional methods:}
Traditional object pose estimation methods use local feature/keypoint matching approaches~\cite{lowe2004distinctive,yu2019unsupervised,tian2019sosnet} to associate 2D images with 3D textured models. Then, the object poses are obtained by solving a Perspective-n-Point (PnP) problem~\cite{lepetit2009epnp}.
However, local feature based methods can only address textured objects. Image template~\cite{hinterstoisser2011multimodal,gu2010discriminative,rios2013discriminatively,zhu2014single}, edge~\cite{hinterstoisser2011gradient,liu2010fast} based methods are developed for pose estimation to tackle textureless objects. When an object undergoes occlusions or drastic illumination changes, those methods might fail to estimate object poses accurately~\cite{peng2019pvnet}.

\noindent\textbf{Fully-supervised deep model based methods:}
Deep learning based methods have demonstrated promising pose estimation performance~\cite{peng2019pvnet,xiang2017posecnn,zakharovdpod,yu6dof,labbe2020cosypose,liu2020leaping}.
Some pioneering approaches, such as Render for CNN~\cite{su2015render}, Viewpoints \& Keypoints~\cite{tulsiani2015viewpoints}, and SSD6D~\cite{kehl2017ssd}, discretize 3D pose space and classify object poses with discrete labels.

Instead of treating pose estimation as a classification task, recent approaches directly regress 3D bounding-boxes~\cite{rad2017bb8,tekin2018real}, local features~\cite{xiang2017posecnn,peng2019pvnet} or coordinate maps~\cite{zakharovdpod,wang2019normalized,Park2019Pix2Pose,zhuang2020lyrn} of objects, and then predict object poses via PnP.
For example, YOLO6D~\cite{tekin2018real} regresses 3D object bounding-boxes from input images. 
BB8~\cite{rad2017bb8} first generates 2D object segmentation masks and then estimates 3D bounding-boxes from the 2D masks.
CPDN~\cite{li2019CDPN}, DPOD~\cite{zakharovdpod} and Pix2Pose~\cite{Park2019Pix2Pose} output the 2D UV coordinates or 3D coordinates of 3D object models from images, while PoseCNN~\cite{xiang2017posecnn} and PVNet~\cite{peng2019pvnet} employ Hough voting to localize object keypoints from estimated vector fields. EPOS~\cite{hodan2020epos} uses symmetric segments of an objects to mitigate the symmetry ambiguity in pose estimation.
Once 2D-3D correspondences are attained, object poses are achieved by solving a PnP problem.

\begin{figure*}[t!]
    \centering\vspace{-1mm}
    \includegraphics[width=0.97\linewidth]{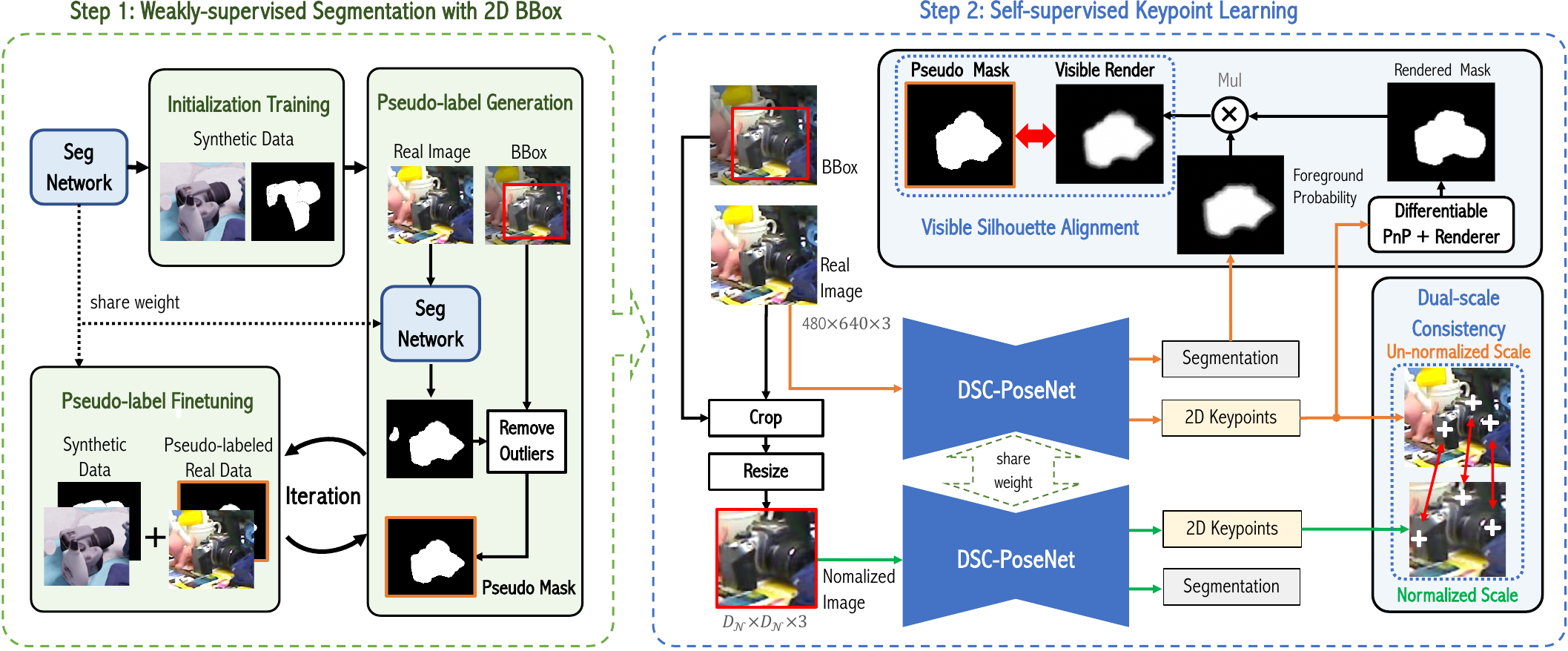}
    \caption{An \textbf{overview} of the training pipeline of DSC-PoseNet. In the first step, we utilize a weakly-supervised segmentation method to generate pseudo masks for real images with only bounding-box (BBox) annotations. In the second step, self-supervised keypoint learning is then developed for training DSC-PoseNet by constructing dual-scale self-supervision signals with a differentiable renderer.}
    \label{fig:overview}
\end{figure*}

\noindent\textbf{Self-supervised deep model based methods:}
Prior arts learn to estimate object poses with ground-truth 3D pose annotations. 
However, when 3D object poses in real images are not available, those methods would suffer severe performance degradation~\cite{wang2020self6d}.
Very recently, some self-supervised methods have been proposed to address the lack of ground-truth pose annotations in real images. Self6D~\cite{wang2020self6d} and DTPE~\cite{rad2018domain} first localize objects and obtain object poses by minimizing the chamfer distances between rendered depth images and real depth images in a self-supervision manner. Although Self6D and DTPE do not require depth images during inference, they need depth images in training. 
\zongxin{AAE~\cite{sundermeyer2018implicit} trains a pose auto-encoder without depth images, but depth information is critical in refining pose results. Without depth information, these methods suffer
dramatic performance degradation.}
NOL~\cite{park2020neural} generates novel views of a given object from a few cluttered real-world images in which object poses are needed. The generated views are self-supervised by the input images. \zongxin{Therefore, it is still an open question of how to estimate object poses only from RGB images for both training and testing.}

\noindent\textbf{Pose refinement:}
Depth images are usually used to refine initially estimated poses. For example, PoseCNN employs Iterative Closest Point (ICP)~\cite{besl1992method} to refine the estimated poses on depth images. 
DeepIM~\cite{li2018deepim} and DPOD~\cite{zakharovdpod} exploit optical flow to modify initial pose estimation by minimizing the differences between the object appearance and the 2D projection of the 3D model. 
Moreover, Hu~\etal~\cite{hu2019single} develop a standalone pose refinement network by removing outliers estimated by other methods and then refining initial pose estimation. 
Cosypose~\cite{labbe2020cosypose} leverages multi-view RGB images to refine each estimated pose in a scene via scene reconstruction.
Considering the vast illumination changes, directly aligning rendered RGB images to input ones for pose refinement might be unreliable. Hence, single RGB image based pose refinement methods require ground-truth poses to bridge the domain gap between rendered images and real ones.
In other words, when the ground-truth pose annotations of real images are not available, single RGB image based methods fail to refine initial pose estimation~\cite{labbe2020cosypose}.

\section{Proposed Method}

\noindent\textbf{Overview.} In this section, we introduce our two-step pose estimation framework with 2D bounding-box (BBox) annotations. As illustrated in Fig.~\ref{fig:overview}, the training paradigm includes (i) weakly-supervised segmentation learning and (ii) self-supervised object keypoint learning.

\subsection{Weakly-supervised Segmentation with BBox}

We aim to obtain object segmentation results from coarse 2D object BBox annotations for real images since segmentation can provide detailed object silhouettes from which object poses can be roughly determined. 
Thus, we design a weakly-supervised segmentation method to generate pixel-level object segmentation. Here, we do not assume the given 2D BBoxes tightly cover object regions, and this further relaxes the labeling requirements.  

Our proposed weakly-supervised segmentation method takes an iterative learning strategy, similar to~\cite{chen2020semi}. As shown in the left of Fig.~\ref{fig:overview}, we first initialize the segmentation network by training it on synthetic data in a fully-supervised way. After initialization, we use the network to generate pseudo-labels for all the BBox-annotated real images and then fine-tune the network on pseudo-labeled real images. We iterate pseudo-label generation and fine-tuning until our segmentation network converges.

\noindent\textbf{Pseudo-label generation and fine-tuning:} 
After initializing the network on synthetic data or fine-tuning on pseudo-labeled real data, we generate (or update) the pseudo segmentation labels for all the real images along with a test-time augmentation, \ie, multi-scale inputs and left-right flips. This augmentation is commonly used in segmentation tasks~\cite{deeplabv3p}. 
Since some pixels in the background might be recognized as foregrounds, we fully exploit 2D BBox annotations to remove those outliers and thus facilitate the network fine-tuning. We further set pixels with low confidence to be uncertain pixels and avoid calculating the segmentation loss for them.

Let $\bm{x}$ and $\bm{y}$ denote an image and a segmentation map respectively, and $y_m \in \{0,1\}$ is the pixel class (background or foreground) at position $m \in \{1,...,M\}$, the uncertain pixel set is defined as,
\begin{equation} \label{equ:uncertain_pixels}
    U_{\bm{x}}=\{m | 1-\sigma < P(y_m=1|\bm{x};\bm{\theta}_{seg}) < \sigma\},
\end{equation}
where $\bm{\theta}_{seg}$ indicates segmentation network parameters, $P$ is a confidence score, and $\sigma$ is a reliable segmentation confidence value. We empirically set $\sigma$ to $0.7$.

\begin{figure*}[t!]
    \centering
    \includegraphics[width=0.65\linewidth]{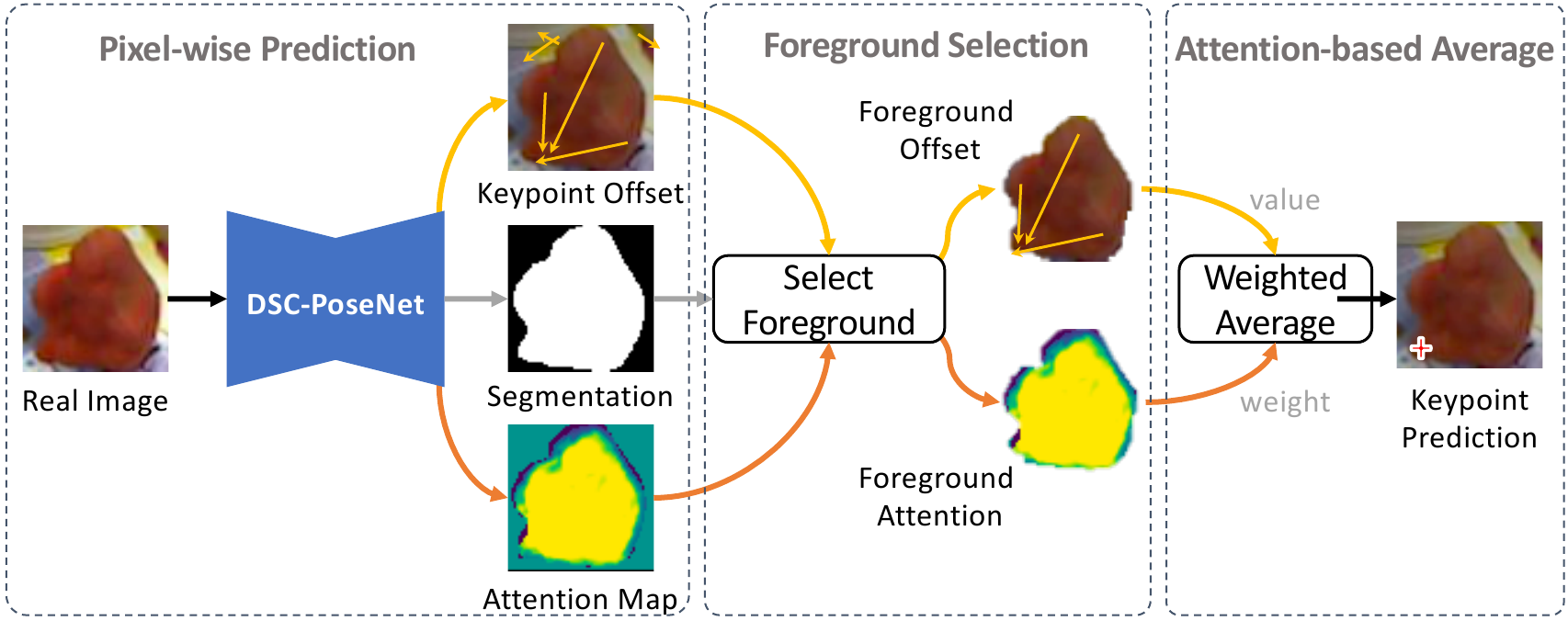}
    \caption{Differentiable 2D Keypoint Coordinate Prediction. We predict the mask of object, the keypoint offset for each pixel and the attention map. The final keypoint prediction is calculated by weighted-averaging the per-pixel predictions in the foreground.}
    \label{fig:keypoint}
\end{figure*}

When fine-tuing on pseudo-labeled real images $\bm{x}_{real}$, the segmentation loss function is,
\begin{equation} \label{equ:seg_loss}
    \mathcal{L}_{seg}=\sum\limits_{m\not\in U_{\bm{x}}} log P(y_m=\hat{y}_m|\bm{x}_{real}, \bm{\theta}_{seg}),
\end{equation}
where $\hat{y}_m$ denotes the pseudo segmentation label.
To prevent the network from fitting to error-prone uncertain regions ($U_{\bm{x}}$), we avoid calculating the loss on them.

The pseudo-label generation and fine-tuning will iterate by $T$ times until the segmentation network converges. Empirically, we found that $T=5$ makes a good balance between training efficiency and performance. The pseudo labels generated after the final iteration will serve as the segmentation labels in the following self-supervised pose estimation step.
Note that some methods, such as Self6D~\cite{wang2020self6d}, do not require object BBoxes in learning segmentation, but they request the photorealism of synthesized images. In contrast, we use coarsely labeled BBoxes to bridge the domain gap between real and synthetic data rather than requiring synthetic data to be similar to real ones.

\subsection{Self-supervised DSC-PoseNet}

Inspired by recent keypoint based estimation methods~\cite{peng2019pvnet,song2020hybridpose}, we use the intermediate object representation, \ie, keypoints, for pose estimation. The final 6DoF poses are obtained by establishing 2D-3D correspondences and solving a PnP problem.
However, the ground-truth positions of keypoints are not available in real images because we do not have 3D pose annotations. Therefore, in the second step of our framework, we design a novel self-supervised dual-scale consistency pose estimation network (DSC-PoseNet) to predict the locations of keypoints and then use the estimated keypoints to predict object poses in real images, as shown at the bottom of Fig~\ref{fig:overview}.

In enforcing dual-scale keypoint consistency, we introduce a normalized scale by normalizing the object scales in real data. In other words, our keypoint network predicts keypoints on both un-normalized and normalized scales. Thus, we can improve the scale robustness by constraining the keypoint consistency between different scales.

To train our pose estimation network, we propose two self-supervised objective functions: visible silhouette alignment (VSA) and dual-scale keypoint consistency (DKC).
VSA measures alignments between the predicted pose's silhouette and the visible silhouette in the image, while DKC evaluates the keypoint estimation consistency in normalized image scale and un-normalized scale.

\noindent\textbf{Differentiable keypoint regression:}
To regress keypoints in a differentiable way, DSNT~\cite{DSNT} averages coordinates based on the weights of a spatial heatmap. However, when a keypoint is occluded, DSNT will fail to predict a valid heatmap. 
PVNet~\cite{peng2019pvnet} predicts a vector field for each keypoint and employs voting to determine keypoint locations. Since the voting procedure is not differentiable, using vector field representation is not applicable to designing a differentiable framework.

To solve all the above issues, we propose to regress 2D keypoint coordinates from all the object pixels (see Fig.~\ref{fig:keypoint}). For each 2D keypoint $k_n ( n=1,...,N$) and each object pixel coordinate $p_m$ at position $m$, DSC-PoseNet will generate an attention weight $a_{nm}$ and predict a keypoint offset $\Delta k_{nm}$. Then, the predicted keypoint $\widetilde{k}_n$ is defined as,
\begin{equation} \label{equ:keypoint}
    \widetilde{k}_n=\frac{\sum\limits_{m\in O}e^{a_{nm}} \widetilde{k}_{nm}}{\sum\limits_{m\in O}e^{a_{nm}}}, ~~~~\widetilde{k}_{nm}=p_m +\Delta k_{nm},
\end{equation}
where $O$ is the foreground segmentation predicted by DSC-PoseNet and $\widetilde{k}_{nm}$ denotes the keypoint prediction from only pixel $p_m$. Compared to the voting process and coordinate heatmap based average, our attention-based average is fully differentiable and can predict occluded keypoints.

For synthetic data $\bm{x}_{syn}$, we can obtain accurate keypoint $k_n$. Thus, we use the smooth $\ell_1$~\cite{girshick2015fast} to learn $k_n$ directly. The keypoint loss function for synthetic data is,
\begin{equation} \label{equ:keypoint}
    \mathcal{L}^{syn}_{key}=\sum\limits_{n=1}^{N}\ell_1(\frac{\widetilde{k}_n(\bm{x}_{syn},\bm{\theta}_{pose}) - k_n(\bm{x}_{syn})}{\sigma S(\bm{x}_{syn})}),
\end{equation}
where $\bm{\theta}_{pose}$ is the parameter of DSC-PoseNet, $\sigma$ is a scale factor, and $S$ is the object scale defined by the length of the longest side of the BBox. 
We observe that the larger an object is in terms of pixels, the greater the variance of the keypoint prediction will be (\ie, pixel deviations). Thus, we employ $S$ to normalize the keypoint estimation errors. We found the normalization with the object scale $S$ is crucial in maintaining a stable training process.

Apart from the keypoint loss, we introduce an auxiliary offset loss,
\begin{equation} \label{equ:offset}
    \mathcal{L}^{syn}_{off}=\sum\limits_{n=1}^{N}\sum\limits_{m\in O}\ell_1(\frac{\widetilde{k}_{nm}(\bm{x}_{syn},\bm{\theta}_{pose}) - k_n(\bm{x}_{syn})}{\sigma S(\bm{x}_{syn})}).
\end{equation}
This loss enforces DSC-PoseNet to regress accurate keypoints from every foreground pixel. Before learning the keypoint regression, we use the offset loss to warm up DSC-PoseNet. This is helpful for initializing the keypoint regression and thus stabilizing the learning of keypoint attention.

\noindent\textbf{Dual-scale keypoint consistency.} 
For real data $\bm{x}_{real}$, we only have pseudo segmentation labels $\widetilde{\bm{y}}$. 
Therefore, we need a self-supervised way to learn keypoints from real data without pose annotations.
Intuitively, the keypoint locations of the same object predicted by our DSC-PoseNet should be consistent across different scales.
Thus, we introduce a dual-scale keypoint prediction consistency constraint for pose estimation, as shown in the bottom right of Fig.~\ref{fig:overview}. 

Given an image $\bm{x}$, we can transform it into a normalized scale $\mathcal{N}(\bm{x})$ by cropping out the object and resizing it to a fixed size, $D_\mathcal{N}\times D_\mathcal{N}$. The 2D keypoints should follow the same transformation,
\begin{equation} \label{equ:dual_scale_cons}
    k_n(\bm{x})=\mathcal{N}^{-1}(k_n(\mathcal{N}(\bm{x}))).
\end{equation}
Such a relation can serve as a dual-scale consistency regularization for our keypoint prediction. 

Apart from the dual-scale consistency, we also utilize an augmentation consistency. Data augmentations (such as random cropping and resizing) are commonly used in pose estimation task for improving the network robustness and relieving over-fitting. The 2D keypoints should be consistent after the augmentation $\mathcal{A}$, \ie,
\begin{equation} \label{equ:aug_cons}
    k_n(\bm{x})=\mathcal{A}^{-1}(k_n(\mathcal{A}(\bm{x}))).
\end{equation}

To further utilize the 3D structure information while reducing prediction errors, we employ BPnP~\cite{bpnp} to compute the object pose from the predicted 2D keypoints, and then re-project the 3D keypoints on a CAD model back to 2D image space using the computed pose. 
To be specific, given a set of keypoint predictions, $\widetilde{\bm{k}}=\{\widetilde{k}_1,...,\widetilde{k}_N\}$, corresponding 3D keypoint set, $\bm{k}^{3D}$, on CAD model, and camera intrinsic matrix, $\mathcal{K}$, the re-projected 2D keypoints $\widetilde{\bm{k}}^{P}$ are,
\begin{equation}\label{equ:reproject}
\begin{aligned}
    \widetilde{\bm{k}}^{P}&=\mathcal{P}(\widetilde{\bm{k}})=\widetilde{R}\bm{k}^{3D}+\widetilde{t},
\end{aligned}
\end{equation}
\begin{equation}
    (\widetilde{R},\widetilde{t})=BPnP(\widetilde{\bm{k}},\bm{k}^{3D},\mathcal{K}),
\end{equation}
where $\widetilde{R}$ and $\widetilde{t}$ are the predicted 3D rotation and translation.
Combining Eqn.~\eqref{equ:dual_scale_cons},~\eqref{equ:aug_cons}, and~\eqref{equ:reproject}, the loss function of our dual-scale consistency for real data is,
\begin{equation}\label{equ:dual}
\begin{aligned}
    \mathcal{L}^{real}_{dual}\!\!&=\!\!\sum\limits_{n=1}^{N}(\ell_1(\frac{\mathcal{A}^{-1}(\widetilde{k}_n(\mathcal{A}(\bm{x}_{real}),\bm{\theta}_{pose})) - \overline{k}_n}{\sigma S(\bm{x}_{real})}) \\
    &+\!\!\ell_1(\frac{\mathcal{A}^{-1}\cdot\mathcal{N}^{-1}(\widetilde{k}_n(\mathcal{N}\cdot\mathcal{A}(\bm{x}_{real}),\bm{\theta}_{pose})) - \overline{k}_n}{\sigma S(\bm{x}_{real})})),
\end{aligned}
\end{equation}
where
\begin{equation}
    \!\!\overline{k}_n\!\!=\!\!\mathcal{P}(\frac{\widetilde{k}_n(\bm{x}_{real},\bm{\theta}_{pose})+\mathcal{N}^{-1}(\widetilde{k}_n(\mathcal{N}(\bm{x}_{real}),\bm{\theta}_{pose}))}{2})
\end{equation}
is the 2D re-projection of the average keypoint prediction over normalized and un-normalized (original) scales of $\bm{x}_{real}$. Similar to Eqn.~\eqref{equ:keypoint}, we use the object scale $S$ to normalize keypoint errors.

\noindent\textbf{Visible silhouette alignment supervision:} 
To estimate object poses from real images, we further optimize DSC-PoseNet by aligning the predicted object pose silhouette with its counterpart in real images.
Given a predicted 3D rotation $\widetilde{R}$, a translation $\widetilde{t}$, the 3D CAD model $\mathcal{M}$, and the camera intrinsic matrix $\mathcal{K}$, we use PyTorch3D~\cite{ravi2020pytorch3d} to render the predicted object mask $\widetilde{\bm{y}}^R=\mathcal{R}(\widetilde{R}, \widetilde{t},\mathcal{M},\mathcal{K})=\{\widetilde{y}^R_1,...,\widetilde{y}^R_M\}$.

We directly use the pseudo labels $\hat{y}$ from our weakly-supervised segmentation step as the object silhouette for real images.
However, the segmentation result only provides visible regions of an object, while the rendered mask covers the entire object structure. In other words,  the alignment will fail when the object is occluded.

To solve the above problem, we use the segmentation branch of DSC-PoseNet to select visible regions of the rendered mask.
Let $\widetilde{y}_m=P(y_m=1|\bm{x};\bm{\theta}_{pose})$ denote the foreground probability predicted by the segmentation branch, the visible rendered mask is,
\begin{equation}
V(\widetilde{\bm{y}}^R)=\{\widetilde{y}^R_m\widetilde{y}_m|m\in{1,...,M}\},
\end{equation}
which is a combination of foreground probability and rendered mask.

After achieving the visible rendered mask, we use the IoU-based Dice loss~\cite{dice} to align it with the pseudo segmentation mask, $\hat{\bm{y}}=\{\hat{y}_1,...,\hat{y}_M\}$, as follows,
\begin{equation} \label{equ:align}
\mathcal{L}^{real}_{align}\!\!=\!\!Dice(\hat{\bm{y}}(\bm{x}_{real},\bm{\theta}_{seg}),V(\widetilde{\bm{y}}^R(\bm{x}_{real},\bm{\theta}_{pose}))),
\end{equation}
\begin{equation}
Dice(\hat{\bm{y}},V(\widetilde{\bm{y}}^R))=1-\frac{\sum\limits_{m=1}^M 2\hat{y}_m\widetilde{y}^R_m\widetilde{y}_m+\epsilon}{\sum\limits_{m=1}^M (\hat{y}_m + \widetilde{y}^R_m\widetilde{y}_m)+\epsilon},
\end{equation}
where $\epsilon$ is a small number for avoiding division by 0.

\begin{table*}[!ht]
\footnotesize
\setlength{\tabcolsep}{3.5pt}
\begin{center}
\begin{tabular}{l| c c c c c c | c c | c c c}
\toprule[1.5pt]
Train data & \multicolumn{8}{|c|}{w/o Real Pose Labels}  & \multicolumn{3}{|c}{with Real Pose Labels} \\
\midrule[0.7pt]
\multirow{2}*{Method} & \multicolumn{6}{|c|}{RGB-based} & \multicolumn{2}{|c|}{RGBD-based} & \multirow{2}*{Tekin~\cite{tekin2018real}} & \multirow{2}*{DPOD~\cite{zakharovdpod}} & \multirow{2}*{CDPN~\cite{li2019CDPN}} \\

 ~ & AAE~\cite{sundermeyer2018implicit} & MHP~\cite{manhardt2018explaining} & Self6D-LB & DPOD~\cite{zakharovdpod} & \textbf{Ours} & \textbf{Ours$^+$} & DTPE~\cite{rad2018domain} & Self6D~\cite{wang2020self6d} & & & \\

\midrule[0.7pt]
Ape & 4.2 & 11.9 & 14.8 & 35.1 & \textbf{35.9} & 31.2 & 19.8 & \textbf{38.9} & 21.6 & 53.3  & \textbf{64.4} \\
Bvise & 22.9 & 66.2 & 68.9 & 59.4 & \textbf{83.1} & 83.0 & 69.0 & \textbf{75.2} & 81.8 & 95.2  & \textbf{97.8} \\
Cam & 32.9 & 22.4 & 17.9 & 15.5 & \textbf{51.5} & 49.6 & \textbf{37.6} & 36.9 & 36.6 & 90.0  & \textbf{91.7} \\
Can & 37.0 & 59.8 & 50.4 & 48.8 & \textbf{61.0} & 56.5 & 42.3 & \textbf{65.6} & 68.8 & 94.1  & \textbf{95.9} \\
Cat & 18.7 & 26.9 & 33.7 & 28.1 & 45.0 & \textbf{57.9} & 35.4 & \textbf{57.9} & 41.8 & 60.4  & \textbf{83.8} \\
Drill & 24.8 & 44.6 & 47.4 & 59.3 & 68.0 & \textbf{73.7} & 54.7 & \textbf{67.0} & 63.5 & \textbf{97.4} & 96.2 \\
Duck & 5.9 & 8.3 & 18.3 & 25.6 & 27.6 & \textbf{31.3} & \textbf{29.4} & 19.6 & 27.2 & 66.0  & \textbf{66.8} \\
Eggbox & 81.0 & 55.7 & 64.8 & 51.2 & 89.2 & \textbf{96.0} & 85.2 & \textbf{99.0} & 69.6 & 99.6  &  \textbf{99.7} \\
Glue & 46.2 & 54.6 & 59.9 & 34.6 & 52.5 & \textbf{63.4} & 77.8 & \textbf{94.1} & 80.0 & 93.8  & \textbf{99.6} \\
HoleP & 18.2 & 15.5 & 5.2 & 17.7 & 26.4 & \textbf{38.8} & \textbf{36.0} & 16.2 & 42.6 & 64.9  & \textbf{85.8} \\
Iron & 35.1 & 60.8 & 68.0 & \textbf{84.7} & 56.3 & 61.9 & 63.1 & \textbf{77.9} & 75.0 & \textbf{99.8} & 97.9 \\
Lamp & 61.2 & - & 35.3 & 45.0 & \textbf{68.7} & 64.7 & \textbf{75.1} & 68.2 & 71.1 & 88.1  & \textbf{97.9} \\
Phone & 36.3 & 34.4 & 36.5 & 20.9 & 46.3 & \textbf{54.4} & 44.8 & \textbf{58.9} & 47.7 & 71.4  &  \textbf{90.8} \\
\midrule[0.7pt]
Mean & 32.6  & 38.8 & 40.1 & 40.5 & 54.7 & \textbf{58.6} & 51.6 & \textbf{59.0}& 56.0 & 82.6  & \textbf{89.9} \\
\bottomrule[1.5pt]
\end{tabular}
\vspace{1mm}
\caption{Comparisons with the state-of-the-art on {\bf LINEMOD} dataset. We present the results for the Average Recall (\%) of ADD(-S). The results except ours and DTPE are copied from~\cite{wang2020self6d}. Ours$^+$: our results trained with additional 10K synthetic data rendered by OpenGL.}\label{tab:linemod}
\end{center}
\end{table*}

The overall self-supervision loss used to train DSC-PoseNet is a combination of all the above losses,
\begin{equation} \label{equ:self}
\mathcal{L}_{self}=\mathcal{L}^{syn}_{key}+\mathcal{L}^{syn}_{off}+\lambda_1\mathcal{L}^{real}_{dual}+\lambda_2\mathcal{L}^{real}_{align},
\end{equation}
where $\lambda_1$ and $\lambda_2$ are the balance factors for $\mathcal{L}^{real}_{dual}$ and $\mathcal{L}^{real}_{align}$, respectively. $\mathcal{L}^{syn}_{key}$, $\mathcal{L}^{syn}_{off}$ and $\mathcal{L}^{real}_{align}$ are employed at both normalized and un-normalized scales.

\section{Experiments}
We conduct experiments on three popular widely-used datasets designed for evaluating 6DOF pose estimation algorithms. We also present a detailed analysis of each key component in our two-step framework. For better understanding, we provide our network only trained on synthetic data as well as real data with ground-truth poses as our lower-bound and upper-bound baselines.

\subsection{Datasets}
\noindent{\bf Synthetic PBR training dataset}~\cite{hodavn2019photorealistic} Our segmentation and pose estimation networks are both trained on synthetic and real images. We employ 50K PBR images rendered with LINEMOD models~\cite{hinterstoisser2012model} as our synthetic data. Note that PBR images exhibit an obvious domain gap with LINEMOD dataset~\cite{hinterstoisser2012model}. 

\noindent{\bf LINEMOD}~\cite{hinterstoisser2012model} consists of 13 objects and each object has around 1,200 images. The CAD models of those objects are also provided. It is regarded as a standard 6DOF pose estimation benchmark. For each object category, 15\% images are used for training and the rest for testing~\cite{rad2017bb8}. LINEMOD exhibits many challenges, such as textureless objects, lighting variations, and cluttered scenes.

\noindent{\bf Occluded LINEMOD}~\cite{brachmann2014learning} provides additional annotations on the bench vise subset in LINEMOD dataset. All the LINEMOD objects are annotated in each image.
As scenes are cluttered, many objects undergo severe occlusion.

\noindent{\bf HomebrewedDB}~\cite{kaskman2019homebreweddb} Following Self-6D~\cite{wang2020self6d}, we also employ HomebrewedDB dataset to test the generalization of our method. Specifically, we adopt the second sequence of HomebrewedDB for evaluation as it contains three objects from LINEMOD (\ie, bench vise, driller, and phone), and those objects are captured in a new environment.

\subsection{Evaluation Metrics}
We employ a commonly used metric, \ie, ADD score~\cite{hinterstoisser2012model}, to evaluate 6DoF pose estimation accuracy.
ADD score measures the average 3D distance between 3D model point clouds transformed by a predicted pose and its corresponding ground-truth. When the distance is less than 10\% of the diameter of a 3D model, the estimated pose is considered correct. 
For symmetric objects, ADD-S~\cite{xiang2017posecnn} is used, where the closest 3D point distance is employed. 

\subsection{Implementation Details}
In our DSC-PoseNet, we choose $8$ keypoints on each object model by the farthest point sampling algorithm. We also apply data augmentation, including random cropping, resizing, rotation, and color jittering, to avoid overfitting. 
We use DeepLabV3+~\cite{deeplabv3p} as the segmentation network in the weakly-supervised step. The backbone of DeepLabV3+ is GCT-ResNet-50~\cite{gct}. To relax the requirements of accurate labeling,
we apply perturbations to ground-truth BBoxes by randomly expanding them up to 15\% of their sizes.

For our DSC-PoseNet, we use ResNet-50~\cite{resnet} as the backbone following Self-6D. 
We set the batch size to $16$ and employ Adam optimizer with an initial learning rate of $1e^{-3}$, which is decayed to $1e^{-5}$ by a factor of $0.85$ every $5$ epochs. The maximum training epoch is set to $120$.
When training on synthetic images, we only use ground-truth keypoints as supervision. When real images are given, we employ the Eqn.~\eqref{equ:dual} and~\eqref{equ:align} to optimize our network. In particular, the resolution of the un-normalized scale is 480$\times$640 pixels, and the normalized one is 64$\times$64 pixels.
During inference, only DSC-PoseNet is used, and the cropping BBoxes for the normalized scale are generated by bounding the object segmentation prediction from the un-normalized scale. Besides, we ensemble the keypoint predictions from both scales by averaging their coordinates.
Our method runs at 7fps on 480$\times$640 images on 1 NVIDIA V100 GPU.

\begin{figure}[t!]
    \centering
    \includegraphics[width=0.9\linewidth]{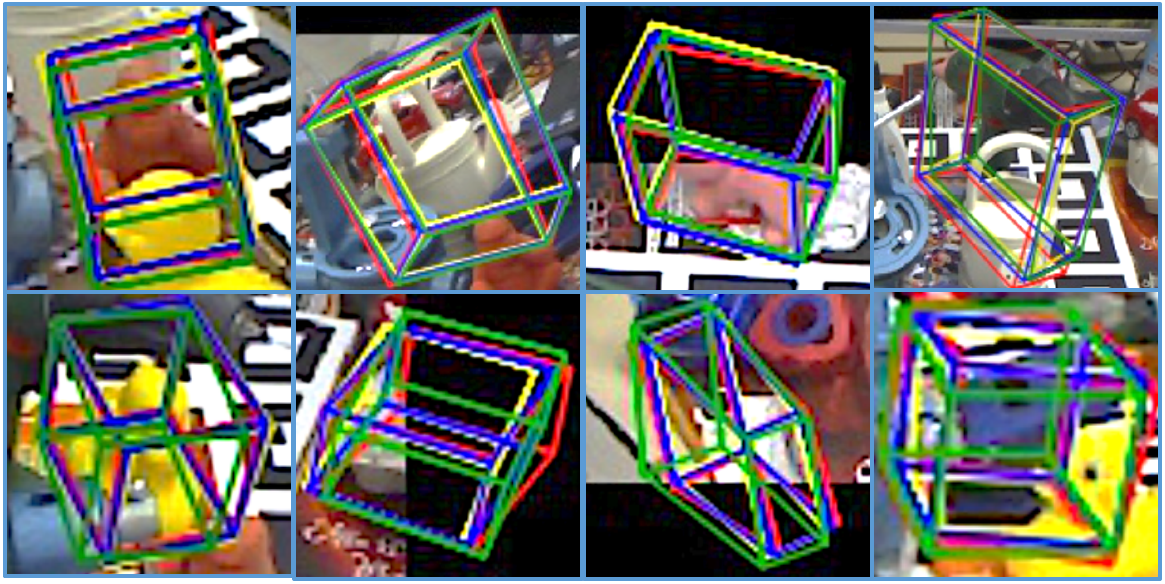}
    \caption{Qualitative results on {\bf Occluded LINEMOD} dataset. \textit{Green}: the ground truth pose. \textit{Red}: un-normalized scale prediction. \textit{Yellow}: normalized scale prediction. \textit{Blue}: ensembled prediction by averaging the keypoints predicted at both the scales.}
    \label{fig:occ_linemod}
\end{figure}

\subsection{Comparison with SOTA}
In this section, we mainly compare our method with state-of-the-art RGB image based methods trained on synthetic datasets, \ie, AAE~\cite{sundermeyer2018implicit}, MHP~\cite{manhardt2018explaining}, DPOD~\cite{zakharovdpod} and PVNet~\cite{peng2019pvnet}, as well as RGBD based methods, \ie, DTPE~\cite{rad2018domain} and Self6D~\cite{wang2020self6d}. Note that, DTPE and Self-6D both use RGBD images during training and only use RGB images in testing. Thus, we categorize Self6D and DTPE as RGBD based methods. Moreover, Self6D also provides an RGB only version, namely Self6D-LB, and we also include it for comparisons.

{\bf Comparisons on LINEMOD:}
As expected, the performance of state-of-the-art methods decreases dramatically on LINEMOD dataset when they are trained on synthetic data, as indicated in Table~\ref{tab:linemod}. This also implies that the domain gap is the main issue rather than different pose representations. 
When depth images are available, RGBD based methods can localize objects and estimate their poses via ICP~\cite{besl1992method}. In this way, they significantly ease the difficulty of pose estimation. As seen in Table~\ref{tab:linemod}, RGBD based methods indeed outperform RGB based methods. 
In contrast, our method even outperforms RGBD based method DTPE and is slightly inferior to Self6D. 
In particular, our method outperforms Self6D on some objects, such as bench vise, cam, and duck, although we did not use depth information during training. We also provide our results with additional synthetic data rendered by OpenGL following Self6D. Under these fairer comparisons, our performance is on par with the results of RGBD based Self6D, as seen in Table~\ref{tab:linemod}.

\begin{table}[!t]
\footnotesize
\begin{center}
\setlength{\tabcolsep}{1.5pt}
\begin{tabular}{l| c c c c c | c }
\toprule[1.5pt]
\multirow{2}*{Method} & \multicolumn{5}{|c|}{RGB-based} & RGBD \\
~ & DPOD~\cite{zakharovdpod} & Self6D-LB & CDPN~\cite{li2019CDPN} & \textbf{Ours} & \textbf{Ours$^+$} & Self6D~\cite{wang2020self6d} \\
\midrule[0.7pt]
Ape & 2.3 & 7.4 & \textbf{20.0} & 13.9  & 9.1 & 13.7 \\
Can & 4.0 & 14.1 & 15.1 & 15.1 & \textbf{21.1} & 43.2 \\
Cat & 1.2 & 7.6 & 16.4 & 19.4 & \textbf{26.0} & 18.7 \\
Drill & 7.2 & 18.0 & 22.2 & \textbf{40.5} & 33.5 & 14.4 \\
Duck & 10.5 & \textbf{12.2} & 5.0 & 6.9 & \textbf{12.2} & 32.5 \\
Eggbox & 4.4 & 18.3 & 36.1 & 38.9 & \textbf{39.4} & 57.8 \\
Glue & 12.9 & 31.4 & 27.9 & 24.0 & \textbf{37.0} & 54.3 \\
HoleP & 7.5 & 11.5 & \textbf{24.0} & 16.3 & 20.4 & 22.0 \\
\midrule[0.7pt]
Mean & 6.3 & 11.5 & 20.8 & 21.9 & \textbf{24.8} & 32.1 \\
\bottomrule[1.5pt]
\end{tabular}
\vspace{1mm}
\caption{Comparisons with the state-of-the-art on {\bf Occluded LINEMOD} dataset.}\label{tab:occ_linemod}
\end{center}
\end{table}
\begin{table}[!t]
\footnotesize
\begin{center}
\setlength{\tabcolsep}{2pt}
\begin{tabular}{l| c c c c | c }
\toprule[1.5pt]
\multirow{2}*{Method} & \multicolumn{4}{|c|}{RGB-based} & RGBD \\
~ & Self6D-Syn & DPOD~\cite{zakharovdpod} & SSD+Ref.~\cite{kehl2017ssd} & \textbf{Ours} & Self6D~\cite{wang2020self6d} \\
\midrule[0.7pt]
Bvise & 37.7 & 52.9 & \textbf{82.0} & 72.9 & 72.1 \\
Drill & 19.2 & 37.8 & 22.9 & \textbf{40.6} & 65.1 \\
Phone & 20.9 & 7.3 & \textbf{24.9} & 18.5 & 41.8 \\
\midrule[0.7pt]
Mean & 25.9 & 32.7 & 43.3 & \textbf{44.0} & 59.7 \\
\bottomrule[1.5pt]
\end{tabular}
\vspace{1mm}
\caption{Comparisons with state-of-the-art on {\bf HomebrewedDB} dataset. Note that we did not finetune our model.}\label{tab:homebrewed}
\end{center}
\end{table}
{\bf Comparisons on Occluded LINEMOD:}
Following previous methods (\ie, DPOD, CDPN and Self6D), we directly test the models trained on the LINEMOD dataset on the BOP~\cite{bop} split of Occluded LINEMOD dataset. 
The comparisons of the state-of-the-art approaches are shown in Table~\ref{tab:occ_linemod}. 
Our method achieves the best overall performance compared to RGB based methods. Note that when Self6D does not use depth images in training, named Self6D-LB, our method even outperforms Self6D-LB. 
Fig.~\ref{fig:occ_linemod} demonstrates our qualitative results on Occlusion LINEMOD.

\begin{table*}[!t]
\footnotesize
\begin{center}
\setlength{\tabcolsep}{4pt}
\begin{tabular}{l|c|c|c|c|c|c|c|c|c|c|c|c|c||c}
\toprule[1.5pt]

~ & Ape & Bvise & Cam & Can & Cat & Drill & Duck & Eggbox & Glue & Holep & Iron & Lamp & Phone & Mean \\
\midrule[0.7pt]
\multicolumn{15}{c}{\textit{Self-supervised Keypoint Learning (ADD metric)}} \\
\midrule[0.7pt]
Ours$^\dagger$ (with Real Pose Labels) & 59.2 & 98.1 & 88.0 & 92.1 & 79.4 & 94.5 & 51.7 & 98.5 & 93.9 & 78.4 & 96.2 & 96.3 & 90.0 & 85.9 \\
\midrule[0.7pt]
\textbf{Ours} & \textbf{35.9} & 83.1 & \textbf{51.5} & \textbf{61.0} & 45.0 & \textbf{68.0} & \textbf{27.6} & \textbf{89.2} & \textbf{52.5} & \textbf{26.4} & 56.3 & \textbf{68.7} & 46.3 & \textbf{54.7} \\
w/o Real RGB images & 23.4 & 75.6 & 11.7 & 40.1 & 26.7 & 53.8 & 14.0 & 73.6 & 26.7 & 19.5 & 56.2 & 39.4 & 20.0 & 37.0 \\
w/o Normalized Scale & 24.7 & 47.0 & 11.9 & 30.4 & 35.7 & 40.1 & 19.5 & 53.7 & 28.7 & 6.4 & 25.0 & 38.8 & 22.7 & 29.6 \\
w/o Keypoint Attention & 10.2 & 60.0 & 26.4 & 33.6 & 29.3 & 39.6 & 8.1 & 43.4 & 40.7 & 8.1 & 16.5 & 45.7 & 32.5 & 30.3 \\
w/o $\mathcal{L}_{real}^{align}$ & 14.8 & \textbf{84.9} & 25.9 & 47.2 & \textbf{47.6} & 56.6 & 13.0 & 88.5 & 40.7 & 8.6 & \textbf{66.0} & 66.3 & \textbf{50.7} & 47.0\\
w/o $\mathcal{L}_{real}^{dual}$ & 32.8 & 76.5 & 42.0 & 49.5 & 47.5 & 58.2 & 21.0 & 69.5 & 51.9 & 11.2 & 51.5 & 63.7 & 41.5 & 47.4 \\
\midrule[0.7pt]
Larger Normalized Scale (128$^2$) & 15.0 & \textbf{85.1} & 40.4 & 50.6 & \textbf{58.6} & \textbf{72.5} & \textbf{35.2} & \textbf{94.0} & \textbf{58.7} & 17.4 & 43.7 & \textbf{79.8} & \textbf{50.4} & 54.0 \\
\midrule[0.7pt]
Un-normalized Scale Predition & 27.9 & 70.6 & 34.9 & 22.2 & 36.2 & 54.0 & 20.4 & 86.8 & 43.5 & 4.9 & 51.4 & 54.4 & 33.9 & 41.6 \\
Normalized Scale Prediction & 22.2 & 63.2 & 46.8 & 11.1 & 37.6 & 51.4 & 20.3 & 80.2 & 31.0 & 6.9 & 37.0 & 40.7 & 35.8 & 37.2 \\
\midrule[0.7pt]
\multicolumn{15}{c}{\textit{Weakly-supervised Segmentation (IoU metric)}} \\
\midrule[0.7pt] 
\textbf{Ours} & \textbf{95.7} & \textbf{92.5} & \textbf{89.4} & \textbf{93.9} & \textbf{92.6} & \textbf{93.9} & \textbf{94.2} & \textbf{95.9} & \textbf{87.5} & \textbf{90.5} & \textbf{90.4} & \textbf{76.2} & \textbf{85.2} & \textbf{90.6} \\
w/o 2D BBox  & 95.2 & 91.9 & 88.7 & 93.4 & 91.8 & 87.5 & 88.4 & 94.9 & 87.1 & 85.8 & 85.9 & 68.8 & 82.0 & 87.8 \\
w/o Real RGB images  & 70.9 & 91.5 & 85.9 & 92.8 & 90.6 & 85.9 & 85.4 & 93.3 & 84.9 & 84.9 & 84.8 & 66.4 & 80.8 & 84.5 \\
\bottomrule[1.5pt]
\end{tabular}
\vspace{1mm}
\caption{Ablation Study on {\bf LINEMOD} dataset. For self-supervised keypoint learning, we report the Average Recall of ADD(-S). For weakly-supervised segmentation, we report the Intersection over Union (IoU) score. }\label{tab:ablation}
\end{center}
\end{table*}
{\bf Comparisons on HomebrewedDB:}
As aforementioned, we employ HomebrewedDB to test the generalization ability of our method. We also compare state-of-the-art method, \ie, DPOD, Self6D, and SSD~\cite{kehl2017ssd} after refinement~\cite{manhardt2018deep} (SSD+Ref.). Note that Self6D re-trains their models on HomebrewedDB and also provides a version only trained on synthetic RGBD data, named Self6D-Syn. As indicated in Table~\ref{tab:homebrewed}, our method still outperforms the methods trained on synthetic data. This also demonstrates that by employing a few 2D annotations on real images, our method significantly minimizes domain gaps and improves the performance of the pose estimation.

Since state-of-the-art RGB based methods do not have mechanisms to use 2D bounding-boxes to train their networks, they may fail to use 2D annotations even though real images are provided. On the contrary, our method is able to leverage 3D and 2D annotations. Therefore, our framework is more generic and more applicable in practice.

\subsection{Ablation Study}
In this section, we analyze the main components of our methods and evaluate the impact of each component on our final pose estimation performance in Table~\ref{tab:ablation}. All the experiments are implemented on LINEMOD. 

{\bf Training data: }
Since we use both synthetic and real data for training, we dissect the contributions of each modality of data. As seen in Table~\ref{tab:ablation}, without real data, our performance degrades significantly and is similar to other RGB based methods. This demonstrates the large domain gap between our synthetic and real data. Moreover, we also incorporate 3D labels of real images to show our method's upper bound, marked as Ours$^\dagger$.

{\bf Attention map: }
We introduce an attention map to compute the weighted average of a keypoint location. For the ablation study, we remove the attention map and directly average all the estimated offsets. Table~\ref{tab:ablation} manifests that our attention map improves the keypoint robustness against the estimation noise.

{\bf Impacts of different losses: }
Our DSC-PoseNet employs a dual-scale strategy to estimate poses. In the training stage, we regularize the keypoint predictions to be consistent for both un-normalized and normalized scales. The network is also optimized by aligning pseudo-labeled and visible rendered masks. Without either of the above self-supervised losses, $\mathcal{L}_{real}^{dual}$ or $\mathcal{L}_{real}^{align}$, the performance of our method decreases significantly, as seen in Table~\ref{tab:ablation}.

{\bf Effects of the normalized scale: }
As shown in Table~\ref{tab:ablation}, the performance of our method degrades significantly without the normalized scale. Moreover, our method is robust for a larger size of the normalized scale (\ie, 128$\times$128), which leads to a similar result. In the inference stage, we ensemble the normalized scale with the un-normalized scale by averaging their keypoint predictions. Table~\ref{tab:ablation} shows that our method achieves much better performance by using ensembled keypoints.

{\bf Importance of 2D annotations: }
2D bounding-box annotations are the only labels for DSC-PoseNet learning. We also investigate how our framework performs when 2D bounding-box annotations are not available. In this case, we employ PBR images to train our segmentation and then generate pseudo labels to finetune the segmentation network iteratively. As expected, some background pixels would be misclassified as the foreground regions, and the IoU score of the segmentation network degrades. Inaccurate segmentation also eliminates the advantage of using real images. Hence, our bounding-box annotations guarantee the performance of pose estimation while reducing domain gaps.

\section{Conclusion}
This paper proposes a novel two-step object pose estimation approach and significantly improves the performance of state-of-the-art RGB based models that are trained without real pose annotations.
Our proposed pose estimation network, dubbed DSC-PoseNet, only uses easily attained 2D bounding-boxes annotations in training.
Thanks to our visible silhouette alignment and dual-scale consistency self-supervision losses, DSC-PoseNet can be trained to estimate object poses without ground-truth supervision and also provides a solution to involve real images when their 3D pose labels are not available. 
Furthermore, although our network is designed for unlabeled real images, it can also be trained with real pose annotations. Hence, our DSC-PoseNet can be adapted to different scenarios, making it more favorable.

\begin{appendices}
\begin{figure*}[t!]
    \centering
    \includegraphics[width=0.95\linewidth]{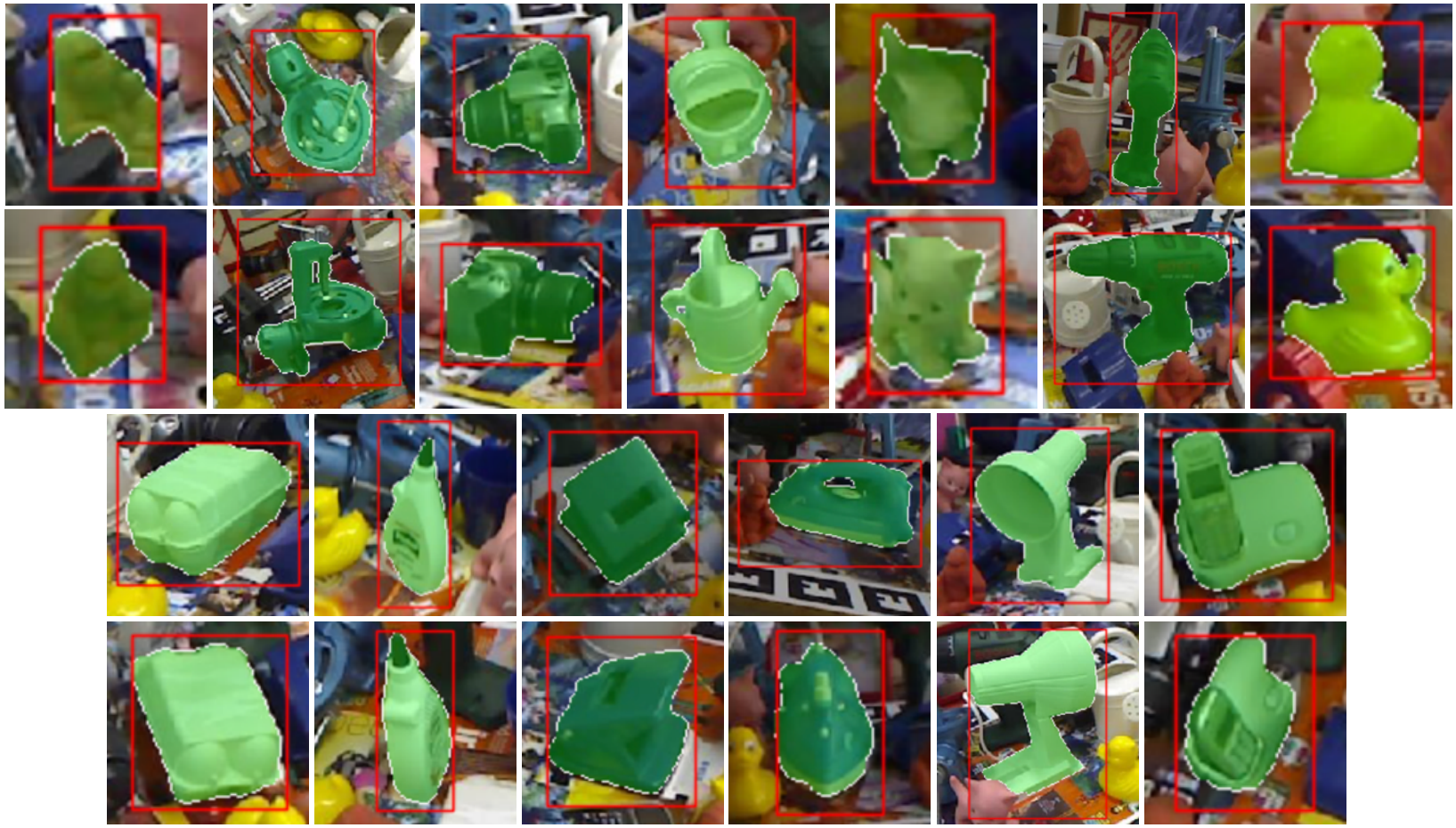}
    \caption{Visualization of given bounding-boxes (\textcolor{red}{red box}) and generated pseudo labels (\textcolor{green}{green region}) on \textbf{LINEMOD}.}
    \label{fig:bbox}
\end{figure*}
\begin{figure}[h!]
    \centering
    \includegraphics[width=0.7\linewidth]{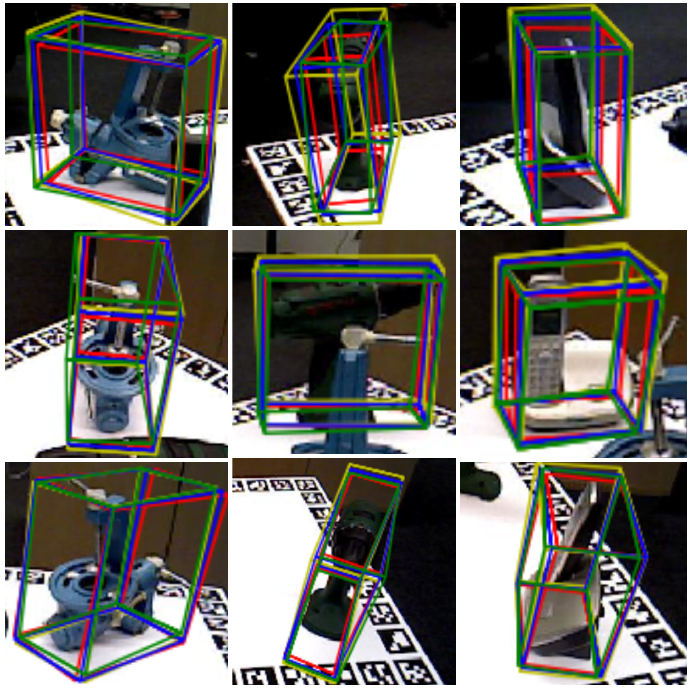}
    \vspace{1mm}
    \caption{Qualitative results on {\bf HomebrewedDB} dataset.}
    \label{fig:hb}
\end{figure}
\section{Bounding-box Visualization}
As shown in Fig.~\ref{fig:bbox}, we do not assume the given 2D bounding-boxes tightly cover object regions, which further relaxes the labeling requirements. For generating loose bounding-boxes, we first extract tight bounding-boxes from the ground-truth object masks. Then we randomly expand the size of tight bounding-boxes by $0\%\sim15\%$ of the width and height.

\begin{table*}[!h]
\centering
\setlength{\tabcolsep}{4pt}
\begin{tabular}{c|ccccccccccccc||c}
\toprule[1.5pt]
$T$ & Ape  & Bvise & Cam  & Can  & Cat  & Driller & Duck & Eggbox & Glue & Holep & Iron & Lamp & Phone & Mean \\
\midrule[0.7pt]
1        & 95.2          & 92.0          & 86.2          & 93.3          & 92.1          & 92.9          & 94.7          & 95.6          & 86.9          & 89.9          & \textbf{90.5} & 72.8          & 83.0          & 89.6          \\
2        & 95.2          & 92.1          & 87.3          & 93.7          & 92.3          & 93.5          & \textbf{95.0} & 95.8          & 87.2          & 90.4          & \textbf{90.5} & 74.4          & 83.9          & 90.1          \\
3        & 95.4          & 92.4          & 88.3          & 93.8          & \textbf{92.7} & 93.7          & 94.8          & \textbf{95.9} & 87.4          & 90.5          & \textbf{90.5} & 75.2          & 84.2          & 90.4          \\
4        & 95.6          & \textbf{92.5} & 88.8          & 93.8          & \textbf{92.7} & \textbf{93.9} & 94.5          & \textbf{95.9} & \textbf{87.5} & \textbf{90.6} & 90.4          & 75.8          & 84.9          & 90.5          \\
\textbf{5 (Ours)} & \textbf{95.7} & \textbf{92.5} & \textbf{89.4} & \textbf{93.9} & 92.6          & \textbf{93.9} & 94.2          & \textbf{95.9} & \textbf{87.5} & 90.5          & 90.4          & \textbf{76.2} & \textbf{85.2} & \textbf{90.6}
\\
\bottomrule[1.5pt]
\end{tabular}
\vspace{2mm}
\caption{The impact of different iteration step numbers ($T$) on our weakly-supervised segmentation on \textbf{LINEMOD}. We evaluate the 
Intersection over Union (IoU) scores between pseudo-labeled masks and ground-truth masks on the training split.}\label{tab:iter}
\end{table*}

\section{Iterative Weakly-supervised Segmentation}
In the weakly-supervised segmentation step, we iterate the pseudo-label generation and fine-tuning process by $T=5$ steps. As seen in Table~\ref{tab:iter}, the quality of pseudo segmentation labels progressively improves as the iteration step increases. The IoU score always converges after $T=3$. 

\section{More Results}
\begin{figure*}[h!]
    \centering
    \includegraphics[width=0.9\linewidth]{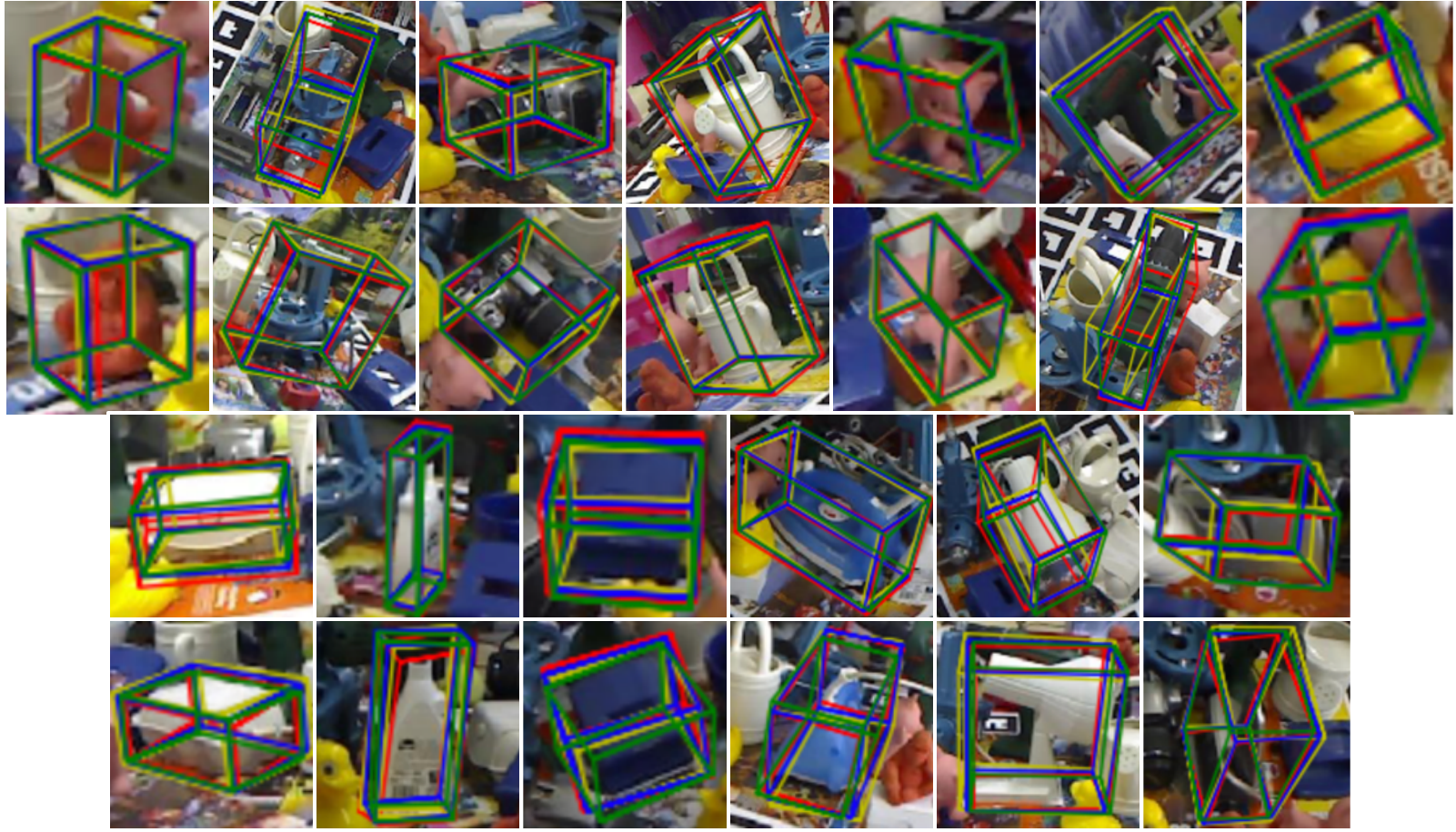}
    \vspace{1mm}
    \caption{Qualitative results on {\bf LINEMOD} dataset. \textit{\textcolor{green}{Green}}: the ground truth pose. \textit{\textcolor{red}{Red}}: un-normalized scale prediction. \textit{\textcolor{yellow}{Yellow}}: normalized scale prediction. \textit{\textcolor{blue}{Blue}}: ensembled prediction by averaging the keypoints predicted at both the scales.}
    \label{fig:linemod}
\end{figure*}
\noindent{\bf Qualitative Results on LINEMOD.} Fig.~\ref{fig:linemod} demonstrates our qualitative results on LINEMOD. The ensembled predictions are always better than the results predicted by a single scale (\ie, either un-normalized scale or normalized scale).

\begin{table}[!h]
\footnotesize
\begin{center}
\setlength{\tabcolsep}{2pt}
\begin{tabular}{l| c c c c | c }
\toprule[1.5pt]
\multirow{2}*{Method} & \multicolumn{4}{|c|}{RGB-based} & RGBD \\
~ & DPOD~\cite{zakharovdpod} & SSD+Ref.~\cite{kehl2017ssd} & \textbf{Ours} & \textbf{Ours$^{FT}$} & Self6D~\cite{wang2020self6d} \\
\midrule[0.7pt]
Bvise & 52.9 & \textbf{82.0} & 72.9 & 78.2 & 72.1 \\
Drill & 37.8 & 22.9 & 40.6 & \textbf{76.2} & 65.1 \\
Phone & 7.3 & 24.9 & 18.5 & \textbf{42.9} & 41.8 \\
\midrule[0.7pt]
Mean & 32.7 & 43.3 & 44.0 & \textbf{65.8} & 59.7 \\
\bottomrule[1.5pt]
\end{tabular}
\vspace{2mm}
\caption{Comparisons with state-of-the-art on {\bf HomebrewedDB} dataset. Ours$^{FT}$: self-supervised fine-tuning by using $15\%$ of real data from HomebrewedDB~\cite{kaskman2019homebreweddb}}\label{tab:hb}
\end{center}
\end{table}

\noindent{\bf Fine-tuning on HomebrewedDB.}
Following Self6D, we provide a self-supervised result by using $15\%$ of real data from HomebrewedDB~\cite{kaskman2019homebreweddb} without pose labels. As shown in Table~\ref{tab:hb}, the self-supervised fine-tuning significantly improves the performance of our RGB based DSC-PoseNet on HomebrewedDB, marked as Ours$^{FT}$. Moreover, Ours$^{FT}$ also outperforms RGBD based Self6D. Fig.~\ref{fig:hb} further demonstrates our qualitative results on HomebrewedDB.

\noindent{\bf Evaluation with BOP~\cite{bop} protocols.}
We also use the Average Recall (AR) used by BOP to evaluate our performance.
The AR scores of our DSC-PoseNet on LINEMOD, OCC-LINEMOD, and HomebrewedDB are 71.0\%, 45.0\%, and 67.4\%, respectively. For comparison, DPOD (synthetic)~\cite{zakharovdpod} achieves an inferior result of 16.9\% on OCC-LINEMOD without using depth and pose annotations.

\end{appendices}

{\small
\bibliographystyle{ieee_fullname}
\bibliography{egbib}
}

\end{document}